\pdfoutput=1

\documentclass[11pt]{article}

\usepackage{style/acl}

\usepackage{times}
\usepackage{latexsym}
\usepackage{xspace}
\usepackage{refstyle}
\usepackage[T1]{fontenc}

\usepackage[utf8]{inputenc}

\usepackage{microtype}

\usepackage{booktabs}
\usepackage{multirow} %
\usepackage{soul}%
\usepackage{tabularx}

\usepackage{amsmath}
\DeclareMathOperator*{\argmax}{argmax} %
\usepackage{pifont}
\newcommand{\cmark}{\ding{51}}%
\newcommand{\xmark}{\ding{55}}%

\usepackage{graphicx}

\usepackage{cleveref}
\crefformat{section}{\S#2#1#3}
\crefformat{subsection}{\S#2#1#3}
\crefformat{subsubsection}{\S#2#1#3}
\crefrangeformat{section}{\S#3#1#4 to~\S#5#2#6}
\crefmultiformat{section}{\S#2#1#3}{ and~\S#2#1#3}{, #2#1#3}{ and~#2#1#3}

\Crefformat{figure}{#2Fig.~#1#3}
\Crefmultiformat{figure}{Figs.~#2#1#3}{ and~#2#1#3}{, #2#1#3}{ and~#2#1#3}
\Crefformat{table}{#2Tab.~#1#3}
\Crefmultiformat{table}{Tabs.~#2#1#3}{ and~#2#1#3}{, #2#1#3}{ and~#2#1#3}
\Crefformat{appendix}{#2Appx.~\S#1#3}
\crefformat{algorithm}{Alg.~#2#1#3}
\Crefformat{equation}{#2Eq.~#1#3}

\newcommand{\stitle}[1]{\vspace{1ex} \noindent{\bf #1.}}

\newcommand{\MODEL}{\mbox{\textsc{Season}}\xspace}

\newcommand{\FIGUREINTRO}{
    \begin{figure}[t]
      \begin{center}
        \includegraphics[width=\columnwidth]{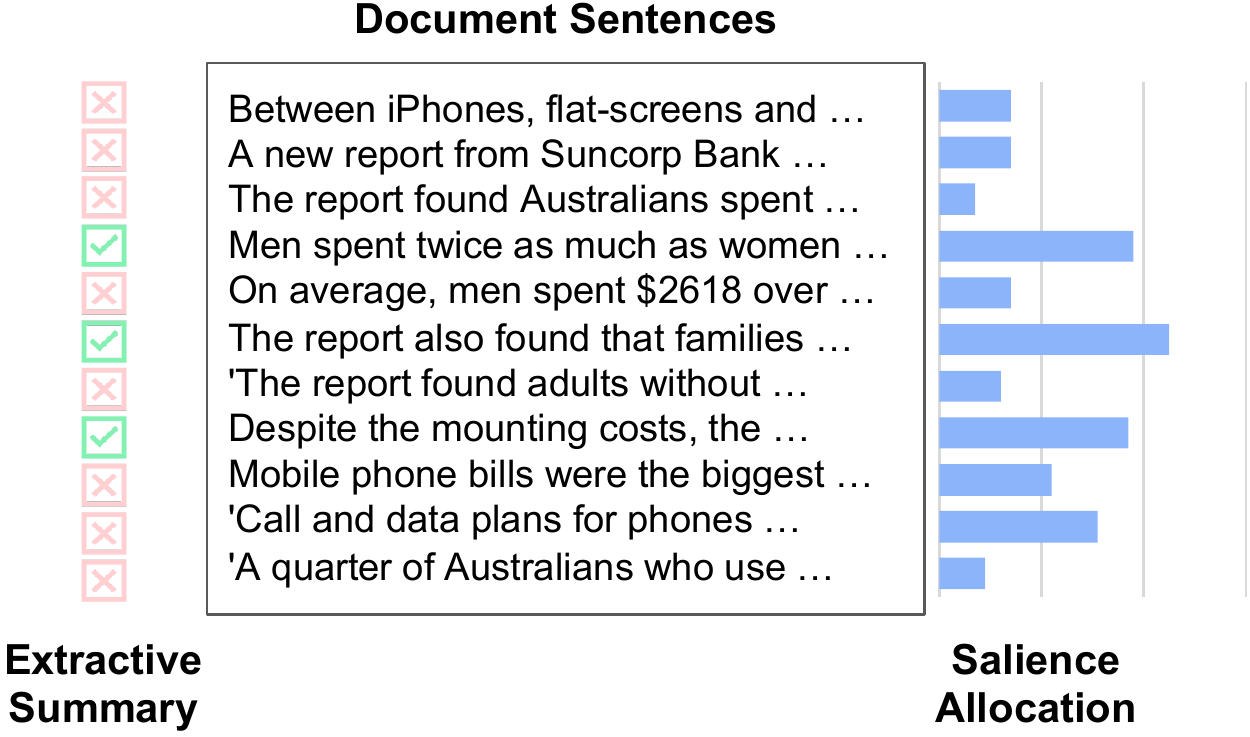} %
      \end{center}
          \caption{Illustration of different guidance. Extractive summary is a strict guidance consisting of extracted sentences labeled with check-mark. Salience allocation is a flexible guidance mapping sentences to different salience degrees shown as a bar chart.}
      
      \label{fig:intro}
      \vspace{-0.5em}
    \end{figure}
}

\newcommand{\FIGUREMODEL}{
    \begin{figure}[t]
      \begin{center}
        \includegraphics[width=\columnwidth]{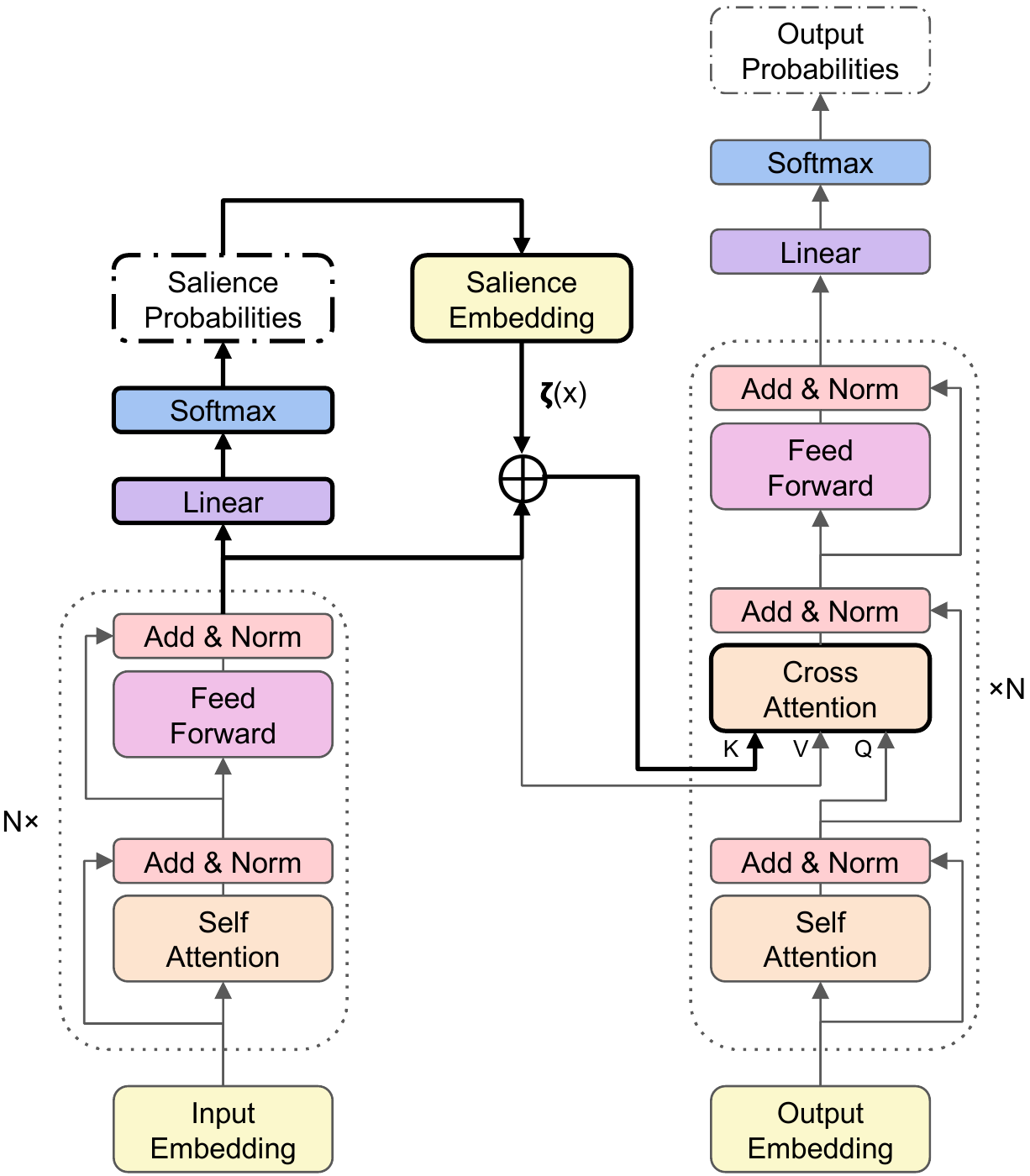} %
      \end{center}
          \caption{Model architecture of \MODEL. The proposed modules are highlighted with bold lines. \MODEL adds a salience predictor on top of the encoder, maps (the expectation of) salience degrees to corresponding embeddings, and adds these salience embeddings to the key vectors of cross attention.}
      
      \label{fig:model}
      \vspace{-0.5em}
    \end{figure}
}

\newcommand{\TABLEMAIN}{
    \begin{table}[t!]
        \centering
        \small
        \begin{tabular}{lccc}
            \toprule 
            \textbf{System} & \textbf{R-1} & \textbf{R-2} & \textbf{R-L}  \\\midrule
            \multicolumn{4}{c}{\textit{\texttt{CNNDM}}} \\\midrule
            LEAD-3 & 40.34 & 17.70 & 36.57 \\
            MatchSum & 44.41 & 20.86 & 40.55 \\
            HAHSum & 44.68 & 21.30 & 40.75 \\ \midrule
            Point-Generator & 39.53 & 17.28 & 36.38 \\
            BART & 44.16 & 21.28 & 40.90 \\
            PEGASUS & 44.17 & 21.47 & 41.11 \\ 
            CIT + SE &45.80 & 22.53 &42.48  \\
            GSum & 45.94 & 22.32 & 42.48 \\\midrule
            BART* & 44.21 &21.23 &41.17 \\
            
            \MODEL &  \underline{\textbf{46.27}} & \underline{\textbf{22.64}} & \underline{\textbf{43.08}} \\
            \midrule %
            \multicolumn{4}{c}{\textit{\texttt{Newsroom}}} \\\midrule
            LEAD-3 & 30.49 & 21.27 & 28.42 \\ \midrule
            Point-Generator & 26.02 & 13.25 & 22.43 \\
            PEGASUS & 45.15 & 33.51 & 41.33\\\midrule
            BART* & 45.50	&33.05	&41.69\\
            \MODEL &  \underline{\textbf{46.00}}	& \underline{\textbf{33.37}}	& \underline{\textbf{42.03}} \\
            \bottomrule
        \end{tabular}
        \caption{\small Results on \texttt{CNNDM} and \texttt{Newsroom} test sets.
        Best scores are in bold.  Scores significantly better than the best baseline model are underlined ($p<0.001$).
        Results with * are reproduced by us.
        Other numbers are from prior papers. 
        }
        \label{tab:main} 
        \vspace{-0.5em}
    \end{table}
}

\newcommand{\TABLESALIENCETHRESHOLDCNNDM}{
    \begin{table}[!t]\centering
        \small
        \setlength{\tabcolsep}{5pt}
        \begin{tabular}{cccc|ccc}\toprule
            \textbf{\#degree} &\textbf{T-1} &\textbf{T-2} &\textbf{T-3} &\textbf{R-1} &\textbf{R-2} &\textbf{R-L}
            \\\midrule
            2 & 85\% & - & - & 45.94 & 22.52 & 42.74 %
            \\
            3 & 50\% &  85\% & - & \textbf{46.38} & \textbf{22.83} & \textbf{43.18}  %
            \\
            4 & 30\% & 50\% &  85\%  & 46.37 & 22.73 & 43.15 %
            \\\bottomrule
        \end{tabular}
        \caption{\small Number of salience degrees, their best percentile thresholds (\texttt{T-x}), and achieved results on CNNDM dev set. %
        }
        \label{tab:threshold}
        \vspace{-0.5em}
    \end{table}
}

\newcommand{\TABLESALIENCEINJECTION}{
    \begin{table}[!t]
        \centering
        \small
        \begin{tabular}{ccccc}\toprule
            \textbf{SACA} & \textbf{MTL} &\textbf{R-1} &\textbf{R-2} &\textbf{R-L} \\\midrule
            \xmark & \xmark  & 44.21 & 21.23 & 41.17 \\
            \xmark & \cmark & 44.57 & 21.55 & 41.49 \\
            \cmark (pred) & \cmark & 46.27 & 22.64 & 43.08 \\
            \cmark (gold) & \cmark & 54.85 & 31.36 & 52.14 \\
            \bottomrule
        \end{tabular}
        \caption{\small Results of models w/ or w/o salience-aware cross-attention (SACA) and multi-task leanring (MTL) on CNNDM test set. For SACA, we provide results with both predicted and gold salience information.}
        \label{tab:ablation_module}
        \vspace{-0.5em}
    \end{table}
}

\newcommand{\TABLEAUXILIARYTASK}{
    \begin{table}[!t]
    \centering
    \small
    \begin{tabular}{cccc}\toprule
        $\alpha$ &\textbf{R-1} &\textbf{R-2} &\textbf{R-L} \\\midrule
        
         0.5 & 46.49 & 22.77 & 43.47 \\
         1.0 & 46.50 & 22.74 & 43.46 \\
         1.5 & 46.48 & 22.81 & 43.44  \\
        \bottomrule
    \end{tabular}
    \caption{\small Results on CNNDM dev set with different loss weights $\alpha$ for salience prediction loss.}
    \label{tab:coefficient}
    \vspace{-1em}
    \end{table}
}

\newcommand{\TABLELABELSMOOTHING}{
    \begin{table}[!t]
        \centering
        \small
        \begin{tabular}{ccccc}\toprule
             \textbf{Strategy} & $\beta$ &\textbf{R-1} &\textbf{R-2} &\textbf{R-L} \\\midrule
            - & -  & 46.38 & 22.83 & 43.18 \\\midrule
            \multirow{2}{*}{all} 
            & 0.1 & 46.47 & 22.71 & 43.42   \\
            & 0.2 & 46.40 & 22.68 & 43.38   \\\midrule
            \multirow{2}{*}{adjacent} 
            & 0.1 & 46.44 & 22.79 & 43.38 \\
            & 0.2 & \textbf{46.48} & 22.81 & \textbf{43.44} \\
            \bottomrule
        \end{tabular}
        \caption{\small Results on CNNDM dev set with different label smoothing strategies.}
        \label{tab:label_smoothing}
        \vspace{-0.5em}
    \end{table}
}

\newcommand{\TABLEMARGINALIZEDDECODING}{
    \begin{table}[!t]
        \centering
        \small
        \begin{tabular}{c|cccc}\toprule
            \textbf{Strategy} & \textbf{$\tau$} &\textbf{R-1} &\textbf{R-2} &\textbf{R-L} \\\midrule
            hard & - & 46.48 & 22.81 & 43.44 \\ \midrule
            \multirow{5}{*}{soft}  
             & 0.1 & 46.76 & 23.08 & 43.57 \\
             & 0.2 & 46.94 & \textbf{23.24} & 43.75 \\
             & 0.5 & \textbf{46.98} & 23.20 & \textbf{43.78} \\
             & 1.0 & 46.59 & 22.75 & 43.44 \\ 
            \bottomrule
        \end{tabular}
        \caption{\small Results on CNNDM dev set with different salience estimation methods. $\tau$ is the sharpening coefficient in softmax.
        }
        \label{tab:salience_expection}
        \vspace{-0.5em}
    \end{table}
}

\newcommand{\FIGURECNNDMSUBSETS}{
    \begin{figure}[t]
      \begin{center}
        \includegraphics[width=0.8\columnwidth]{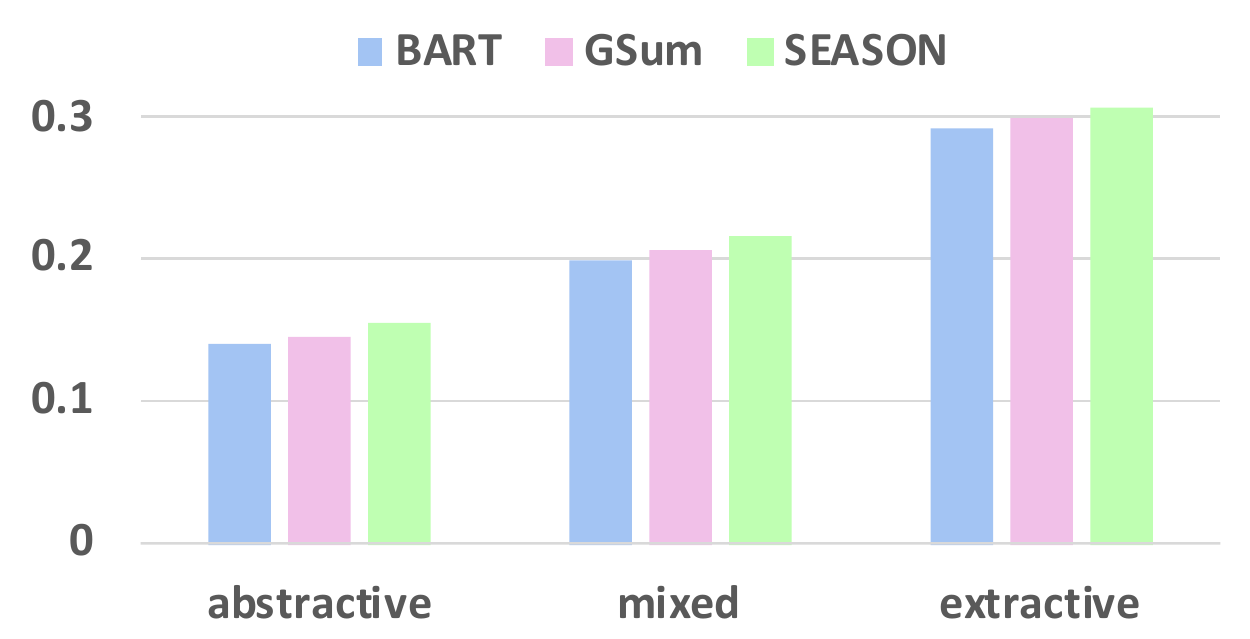} %
      \end{center}
      \caption{\small R-2 scores by density on CNNDM test set.
      }
      \label{fig:cnndm_subsets}
      \vspace{-0.5em}
    \end{figure}
}

\newcommand{\TABLEHUMANEVAL}{
    \begin{table}[!t]
        \centering
        \small
        \setlength{\tabcolsep}{3.5pt}
        \begin{tabular}{lccc}\toprule
             &\textbf{Informativeness} &\textbf{Faithfulness} &\textbf{Fluency} \\\midrule
            BART   & 87.21 & 76.77 & 85.86 \\
            GSum   & 78.45 & \textbf{79.46} & 26.94*\\
            \MODEL & \textbf{88.89} & 78.11 & \textbf{87.88} \\
            Ground-Truth   & 77.78 & 75.76 & 72.39	\\\bottomrule
        \end{tabular}
        \caption{\small Percentage of positive votes ('Yes') on informativeness, faithfulness and fluency of summaries. *GSum predictions provided by the authors are lower-cased and lemmatized, which hinders the fluency. }
        \label{tab:human_evaluation_questions}
        \vspace{-0.5em}
    \end{table}
}

\newcommand{\TABLEHUMANRANK}{
    \begin{table}[!t]
        \centering
        \small
        \begin{tabular}{lccccc}\toprule
             &\textbf{1st} &\textbf{2nd} &\textbf{3rd} & \textbf{4th} & \textbf{avg.}\\\midrule
            BART   & 34.68 & 30.64 & 21.21 & 13.47 & 2.13 \\
            GSum   & 11.11 & 15.49 & 24.92 & 48.48 & 3.11 \\
            \MODEL & \textbf{35.02} & 29.63 & 24.24 & \textbf{11.11} & \textbf{2.11} \\
            Ground-Truth   & 19.19 & 24.24 & 29.63 & 26.94 & 2.64	\\\bottomrule
        \end{tabular}
        \caption{\small Percentage of ranking and the average rank by human evaluation.}
        \label{tab:human_evaluation_ranking}
        \vspace{-1em}
    \end{table}
}

\newcommand{\TABLECASE}{
    \begin{table*}[!t]
        \centering
        \scriptsize
        \begin{tabular}{p{0.00\textwidth} p{0.05\textwidth}  p{0.9\textwidth}}\toprule
            & \textbf{BART} & \textcolor{blue}{New York-based writer Danielle Page set out to ask every cabbie she came across to} dispense their best piece of relationship advice. The drivers, many of whom are married themselves, revealed their personal tips, life lessons and cultural anecdotes all in the name of love. \\
            & \textbf{GSum} & \textcolor{blue}{New York-based writer Danielle Page set out to ask every cabbie she came across to} dispense their best piece of relationship advice in hopes of unlocking the key to a successful union. The drivers, many of whom are married themselves, revealed their personal tips, life lessons and cultural anecdotes all in the name of love. A 60-year-old named Michael revealed that his trick to marital bliss is putting his wife's happiness above his own -- but insists that what really makes a relationship work is finding a partner who will do the same for you. \\
            & \textbf{\MODEL} & \textcolor{blue}{New York-based writer Danielle Page set out to ask every cabbie she came across to} dispense their best piece of relationship advice. \\
            & \textbf{Gold} & New York-based writer Danielle Page set out to ask every cabbie she came across to \textcolor{red}{share their tips on finding - and keeping - a partner.} \\
            \midrule
            & \textbf{BART} & Researchers from Texas A\&M School of Public Health \textcolor{blue}{found that hospitalizations from car crashes dropped 7 percent between 2003 and 2010 in the 45 states with texting bans. Arizona, Texas, Montana, Missouri, and Oklahoma are the only five states in America that do not have texting at the wheel bans for all drivers.}\\
            & \textbf{GSum} & Researchers from Texas A\&M School of Public Health \textcolor{blue}{found that hospitalizations from car crashes dropped 7 percent between 2003 and 2010 in the 45 states with texting bans when compared to states with no restrictions.} Drivers between the ages of 25 and 40 are the most likely group of people to get in an accident related to texting and driving.\\
            & \textbf{\MODEL} & Researchers from Texas A\&M School of Public Health \textcolor{blue}{found that hospitalizations from car crashes dropped 7 percent between 2003 and 2010 in the 45 states with texting bans. Arizona, Texas, Montana, Missouri, and Oklahoma are the only five states in America that do not have texting at the wheel bans for all drivers. The study found that older drivers were more likely to make a texting and driving mistake than a younger driver.} \\ 
            & \textbf{Gold} & Study found that hospitalizations from car crashes dropped 7 percent between 2003 and 2010 in the 45 states with texting bans. Arizona, Texas, Montana, Missouri, and Oklahoma are the only five states in America that do not have texting at the wheel bans for all drivers. The study also found that older drivers were more likely to make a texting and driving mistake than a younger driver. \\
            \bottomrule
        \end{tabular}
        \caption{\small Case Study. Continuous word spans overlapped with the gold summary of more than 3 words are in \textcolor{blue}{blue}. Continuous word spans in the gold summary not covered by any prediction are in \textcolor{red}{red}. Baselines may suffer from extra details or information loss due to no or imperfect salience guidance.
        }
        \label{tab:case}
        \vspace{-1em}
    \end{table*}
}

\title{Salience Allocation as Guidance for Abstractive Summarization}

\author{
  Fei Wang$^{\dagger*}$, Kaiqiang Song$^{\ddagger*}$, Hongming Zhang$^{\ddagger}$, Lifeng Jin$^{\ddagger}$, Sangwoo Cho$^{\ddagger}$ \\ 
  \textbf{Wenlin Yao$^{\ddagger}$, Xiaoyang Wang$^{\ddagger}$,  Muhao Chen$^{\dagger}$ and Dong Yu$^{\ddagger}$}
 \\
  $^\dagger$University of Southern California;\; $^\ddagger$Tecent AI Lab, Seattle\\
  \small \texttt{\{fwang598,muhaoche\}@usc.edu} \\ 
  \small \texttt{\{riversong,hongmzhang,lifengjin,swcho,wenlinyao,shawnxywang,dyu\}@global.tencent.com}}

\begin{document}
\maketitle

\renewcommand{\thefootnote}{\fnsymbol{footnote}}
\footnotetext[1]{Work done during Fei Wang's internship at Tencent AI Lab Seattle. The first two authors contributed equally.}
\renewcommand{\thefootnote}{\arabic{footnote}}

\begin{abstract}

Abstractive summarization models typically learn to capture the salient information from scratch implicitly.
Recent literature adds extractive summaries as guidance for abstractive summarization models to provide hints of salient content and achieves better performance.
However, extractive summaries as guidance could be over strict, leading to information loss or noisy signals.
Furthermore, it cannot easily adapt to documents with various abstractiveness.
As the number and allocation of salience content pieces varies, it is hard to find a fixed threshold deciding which content should be included in the guidance.
In this paper, we propose a novel summarization approach with a flexible and reliable salience guidance, namely \MODEL ({\underline{S}}alienc{\underline{E}} {\underline{A}}llocation as Guidance for Abstractive {\underline{S}}ummarizati{\underline{ON}}).
\MODEL utilizes the allocation of salience expectation to guide abstractive summarization and adapts well to articles in different abstractiveness.
Automatic and human evaluations on two benchmark datasets show that the proposed method is effective and reliable.
Empirical results on more than one million news articles demonstrate a natural fifteen-fifty salience split for news article sentences, providing a useful insight for composing news articles.\footnote{Code and model weights are available at \url{https://github.com/tencent-ailab/season}.}

\end{abstract}

\section{Introduction}

Abstractive summarization %
seeks to generate %
concise descriptions about synoptic information of longer documents
\cite{rush2015neural,nallapati2016abstractive,see2017get}. Tackling this task can provide users with improved dissemination and acquisition of more readable content in long documents.
More concretely, it allows for enhanced selection, compression and retrieval of Web-scale textual information that benefits other NLP tasks such as machine reading comprehension \cite{inoue-etal-2021-summarize}, mention linking \cite{cheng2015summarizing}, claim verification \cite{yin-etal-2021-docnli}, and information extraction \cite{lu2022summarization}.

Abstractive summarization models are typically trained end-to-end using large collections of paired corpora of raw documents and human-written summaries to directly perform sequence-to-sequence generation.
In terms of deciding what to include in the generated summaries, these models %
implicitly learn to capture the salient information from scratch.
Accordingly, recent literature has attempted to add auxiliary extractive salience guidance for abstractive summarization models to give them a higher-level understanding of input documents,
among which, %
extractive summaries appear to provide the most effective guidance
\cite{li2020keywords,jin2020semsum,dou2021gsum}.
Methods following this strategy learn to first perform extractive summarization, then perform abstraction on top of the extractive summaries \cite{hsu2018unified,pilault2020extractive,dou2021gsum}.

\FIGUREINTRO

However, incorporating extractive summaries as a form of guidance is evidently imperfect, %
even though it improves the overall performance of abstractive summarization in some cases \cite{dou2021gsum}: %
1) Extractive summaries are not reliable guidance. When there are too many summary-worthy sentences in the document, selecting a part of them may prone to information loss.
When there are too few or no summary-worthy sentences, using the selected extractive summaries could be noisy and confusing to the model.
2) Extractive summaries are not flexible to adapt to different cases.
The number and allocation of salience content pieces can vary by documents. Rather than extracting a fixed number of sentences, a flexible guidance should select salient content based on document properties.
An imperfect selection process may also lead to further model biases, such as positional biases or length biases \cite{zhong-etal-2019-searching}.
As the summarization process can differ %
for distinct documents \cite{grusky2018newsroom,koupaee2018wikihow}, %
a reliable guidance should allow flexible content selection, and be adaptive to documents with different abstractiveness.

In this paper, we propose a novel summarization approach with a flexible and reliable salience guidance,
namely \MODEL ({\underline{S}}alienc{\underline{E}} {\underline{A}}llocation as Guidance for Abstractive {\underline{S}}ummarizati{\underline{ON}}). 
Salience is the degree to which a sentence contributes to the central idea of a document, and its allocation means how salience is distributed among all sentences in a document.
To estimate the salience allocation,
a linear classifier is trained on top of the encoder.
This estimation is incorporated into the decoder %
with Salience-Aware Cross-Attention (SACA).
It provides the flexibility to decide how much signal to accept from the salience guidance to supervise the abstractive summarization.
The ground-truth salience label is assigned to each sentence based on its similarity with the ground-truth summary.
Meanwhile, the number of salience degrees and their cut-off thresholds are decided based on the corpus to balance informativeness and prediction accuracy.
To further improve the robustness of the summarization model, 
we apply label smoothing between adjacent salience degrees during training,
and use the expectation of salience as a more robust salience estimation.

The technical contributions of this work are three-fold.
    \emph{First}, we develop a new method for abstractive summarization on Transformer-based encoder-decoder architecture with the allocation of salience expectation as flexible guidance (\Cref{sec:method}).
    Our method 
    provides reliable guidance that adapts well to articles in different abstractiveness (\Cref{sec:abstractiveness}).
    \emph{Second}, we show the effectiveness and reliability of our proposed method comparing to the existing methods in both automatic (\Cref{sec:main_result}) and human evaluation  (\Cref{sec:human}).
    \emph{Third}, empirical results on more than one million news articles show a natural \emph{fifteen-fifty salience split} for news article sentences (\Cref{sec:principle}),
    providing a useful insight for composing news articles.

\section{Related Work}

\stitle{Joint extractive and abstractive summarization}
Extractive summarization and abstractive summarization are two general paradigms of text summarization \cite{see2017get,grusky2018newsroom}. Extractive summarization ensures the faithfulness of the generated summary but is not able to properly summarize documents when rephrasing is needed \cite{liu2009extractive}. Abstractive summarization, comparatively, is more flexible but may suffer from hallucination \cite{maynez2020faithfulness}.

A series of studies attempt to benefit from the advantages of both paradigms by combining them. \citet{hsu2018unified} encourage the word-level attention of an abstractive summarization model and the relative sentence-level extraction probability from an extractive summarization model to be consistent. More recent studies show that conducting abstractive summarization with extractive summaries as a part of the input leads to better performance \cite{saito2020abstractive,pilault2020extractive,dou2021gsum}. 
Extractive summarization can also work as an effective content selector for abstractive summarization when summarizing long documents \cite{manakul2021long}. 
Some studies~\cite{gehrmann2018bottom,li2020keywords,saito2020abstractive} also consider to extract key words or phrases instead of summary worthy sentences as guidance, but their performances are not as good as those using sentences \cite{dou2021gsum}.

Our work extends the strict extractive summary guidance to a soft guidance of salience allocation. The proposed guidance is more flexible, reliable and adaptive, leading to better performance.  %

\stitle{Selective attention}
Selective attention is a psychological concept referring to the differential processing of simultaneous sources of information \cite{johnston1986selective}.
Incorporating prior knowledge through selective attention is widely explored in natural language processing, especially in recent NLP models with attention mechanism \cite{lin2016neural,sukhbaatar2019adaptive,pruthi2020learning,Beltagy2020Longformer,wang2022robust}.
To modify the summarization process with selective attention,
previous studies either adjust the attention scores based on content selection probabilities directly \cite{hsu2018unified,saito2020abstractive,li2021ease}, or appending selected content in the input \cite{saito2020abstractive,dou2021gsum}.
Recent studies show that the latter method with sentence-level content selection performs better \cite{dou2021gsum}.

Different from prior studies, \MODEL maps salience degrees to distinct %
embeddings and adds them to the encoder outputs as key vector for cross-attention.
This gives our model the flexibility to decide how much signal to accept from the salience guidance for supervising the abstractive summarization process.
This strategy achieves better performance 
in comparison with previous salience-guided selective attention methods.

\section{\MODEL}\label{sec:method}
\FIGUREMODEL

In this work, we employ a Transformer-based encoder-decoder model for abstractive summarization.
As shown in \Cref{fig:model}, our model \MODEL encapsulates salience prediction and text summarization in a single network.
We perform multi-task end-to-end training, and inference via one forward pass.
During training, the model jointly learns to predict the degree of salience for each sentence and is guided with ROUGE-based ground-truth salience allocation to generate the abstractive summary. 
During inference, \MODEL predicts the expected salience allocation intermediately with the encoder outputs, and uses this predicted information to guide the decoder to generate the summary.

\subsection{Problem Formulation}
Our assumption comes from an intuition that knowing the content salience allocation helps %
the model to pay attention to important content and generate more informative summaries.
Although the content salience allocation is a built-in attribute of the source document, it is hard for the model to leverage this attribute without direct supervision~\cite{li2020keywords,saito2020abstractive,dou2021gsum}.

Let $\mathbf{x}$ be the sequence of input tokens in the source document, and $\mathbf{y}$ be the sequence of the summary tokens, where every token $x_i$ or $y_i$ is in the vocabulary $\mathbf{\mathcal{V}}$.
We use $z_j$, where $j \in \{1, \dots, N\}$, to represent the salience degree of the $j$-th sentence in the input document.
We define $o_{i}$ as the sentence index for the $i$-th token, where $o_{i} \in \{1, \dots, N\}$.
The salience allocation is defined as $\mathbf{\zeta}(\mathbf{x}) =[\mathbf{f}(z_{o_1}), \dots, \mathbf{f}(z_{o_{|\mathbf{x}|}})]$.\footnote{$\mathbf{f}(\cdot)$ is a function that maps the sentence salience degree to an embedding vector. In our implementation, we use the ground-truth salience embedding for training, and the expected embedding over the inferred salience distribution for testing.}
The problem can be formulated as follows:
\vspace{-1mm}
\begin{equation}
    P(\mathbf{y}|\mathbf{x}) = \prod_{k=1}^{|\mathbf{y}|}{p_{\theta}(y_{k}|\mathbf{y}_{<k}, \mathbf{x}, \mathbf{\zeta}(\mathbf{x}))} .
    \label{eq:formulation}
    \vspace{-1mm}
\end{equation}
In \Cref{eq:formulation}, each token prediction is conditioned on the previously decoded summary tokens, the input tokens in the source document, and the allocation of salience of the source document.

\subsection{Salience Allocation Prediction}
To predict salience degrees of input sentences, we slightly modify the encoder input sequence by adding a special token at the beginning of each sentence, obtaining their last-layer hidden states as sentence representations:
\vspace{-1mm}
\begin{equation}
    [\mathbf{h}^{sent}_1, \dots, \mathbf{h}^{sent}_n] = Encoder(\mathbf{\hat{x}}) ,
    \label{eq:sentence_encoder_hidden}
    \vspace{-1mm}
\end{equation}
where $\mathbf{h}^{sent}_j, j \in \{1, \dots, N\}$, is the contextualized embedding of the $j$-th sentence, and $\mathbf{\hat{x}}$ is the modified input sequence.
Then, sentence representations are fed into a single-layer classification head:
\vspace{-1mm}
\begin{equation}
    P(z_j=l|\mathbf{x}) \propto \exp(\frac{\mathbf{w}_l^{T}\mathbf{h}^{sent}_j + b_{u}}{\tau}) ,
    \label{eq:cls_prob}
    \vspace{-1mm}
\end{equation}
where $\tau$ is a sharpening coefficient for the salience degree distribution, $l \in \{1, \dots, L\}$ is the index of salience degree, $L$ is the number of salience degrees, $\mathbf{w}_l$ and $b_{l}$ are trainable parameters. 
We provide discussions on $L$ and $\tau$ in \Cref{sec:principle} and \Cref{sec:ablation} respectively.
The design above allows the model predict salience allocation with minimal modifications on the architecture.

\subsection{Salience-Aware Cross-Attention}
To explicitly incorporate the salience allocation into the model , we develop a salience-aware cross-attention (SACA) module.
SACA first maps the salience degrees to trainable salience embeddings:  %
\vspace{-1mm}
\begin{equation}
    \mathbf{f}(z_j) = \textbf{Emb}(z_j) .
    \vspace{-1mm}
\end{equation}
This operation is intuitive when using ground-truth salience degrees.
For predicted salience degrees, SACA needs to perform an estimation on the salience embedding with the inferred salience distribution.
A simple \textbf{hard} estimation can be achieved by
directly taking the embedding of degree $l$ that maximizes the probability:
\vspace{-1mm}
\begin{equation}
    \mathbf{f}(z_j) = \textbf{Emb}(\argmax_{l} P(z_j=l|\mathbf{x})) .
    \label{eq:function_f_hard}
    \vspace{-1mm}
\end{equation}
However, this direct estimation does not take the uncertainty of prediction into consideration, so we propose the \textbf{soft} estimation that calculates the expectation for the salience embedding:
\vspace{-1mm}
\begin{equation}
    \mathbf{f}(z_j) = \sum_{l=1}^{L}{\textbf{Emb}(z_j=l)P(z_j=l|\mathbf{x})} .
    \label{eq:function_f_soft}
    \vspace{-1mm}
\end{equation}
We compare these two estimation methods comprehensively in \Cref{sec:ablation}.
Next, SACA incorporates the salience allocation in the cross-attention layer to guide summary generation on the decoder side.
SACA adds the sentence salience embedding to the encoder hidden state of each token belonging to the sentence as the key state for cross-attention.
The cross-attention is formulated as:
\vspace{-1mm}
\begin{equation*}
\text{CrossAttn}(Q, K, V) = \text{MultiheadAttn}(Q, K, V) ,
\end{equation*}
where the attention query $Q = \textbf{h}^{decoder}$ thereof corresponds to the hidden state of the decoder, the attention key $K = \textbf{h}^{encoder} + \mathbf{\zeta}(\mathbf{x})$ is the sum of the encoder hidden state and the salience embedding, and
the value $V = \textbf{h}^{encoder}$ is composed of the original encoder hidden state.
In comparison with adding salience scores to cross-attention scores directly, SACA allows the model to learn how much signal to take from the salience guidance.

\subsection{Learning Objectives}
In training, \MODEL learns to predict the salience allocation and generate the summary simultaneously.
For salience prediction, we use the averaged cross-entropy loss on each predicted sentence:
\vspace{-0.1mm}
\begin{equation}
    \mathcal{L}_{cls}= -\frac{1}{N}\sum_{j=1}^{N}{\log{P(z_j|\mathbf{x})}} .
    \label{eq:loss_cls}
    \vspace{-1mm}
\end{equation}
In addition, we apply label smoothing~\cite{diaz2019soft} to the salience degrees for denoising.
Specifically, a probability $\beta$ is evenly assigned to salience degrees adjacent to the ground-truth degree.
Analysis in \Cref{sec:ablation} shows its effectiveness comparing with common label smoothing.
For summary generation, we use the ground-truth salience allocation as input, and apply the averaged cross-entropy loss on each predicted token as below:
\vspace{-0.1mm}
\begin{equation}
    \mathcal{L}_{lm}= -\frac{1}{|\mathbf{y}|}\sum_{k=1}^{|\mathbf{y}|}{\log{p_{\theta}(y_{k}|\mathbf{y}_{<k}, \mathbf{x}, \mathbf{\zeta}(\mathbf{x}))}} .
    \label{eq:loss_lm}
    \vspace{-1mm}
\end{equation}
We further combine two loss functions together with a coefficient $\alpha$ that balances the two:
\vspace{-0.1mm}
\begin{equation}
    \mathcal{L}_{total} = \mathcal{L}_{lm} + \alpha \mathcal{L}_{cls} .
    \label{eq:loss_total}
    \vspace{-1mm}
\end{equation}
\Cref{sec:ablation} shows that \MODEL is not sensitive to $\alpha$.

\section{Experiment}
In this section, we first describe our experimental setting, %
including datasets, baselines, evaluation metrics and implementation details (\Cref{sec:setup}).
Then, we show the model performance on two summarization datasets (\Cref{sec:main_result}), and provide an insight on salience threshold selection (\Cref{sec:principle}).

\subsection{Experimental Setup}\label{sec:setup}
\stitle{Datasets} 
We evaluate our method on two news summarization datasets. For both datasets, we use the original news article as input and the human-written summary as the ground-truth output.
\textit{CNNDM}~\cite{see2017get} consists of news articles and their human-written abstracts from CNN and Daily Mail websites, including 287,226/13,368/11,490 training/validation/test pairs.
On average, each article has 781 words and each abstract contains 56 words.
\textit{Newsroom}~\cite{grusky2018newsroom} contains news articles and summaries written by authors and editors from 38 major newsrooms published between 1998 and 2017. The dataset includes 995,041/108,837/108,862 training/validation/test pairs. On average, each article has 659 words and each summary has 27 words.

\stitle{Metrics}
We report widely used ROUGE metrics~\cite{lin2004rouge}, %
including ROUGE-1 (R-1), ROUGE-2 (R-2), and sentence-level ROUGE-L (R-L) F$_1$ scores with rouge-score python package.\footnote{\url{https://pypi.org/project/rouge-score/}}

\stitle{Baselines}
We compare our system with three types of strong baselines, including extractive, abstractive, and mixed %
summarization methods.
\textit{LEAD-3}~\cite{see2017get} is a common extractive summarization baseline %
that extracts the first three sentences as the document summary.
\textit{BertSum-Ext}~\cite{liu2019fine} fine-tunes BERT for extractive summarization.
\textit{MatchSum}~\cite{zhong2020extractive} learns the semantic matching between candidate summary and source document using contrastive learning.
\textit{HAHSum}~\cite{jia2020neural} incorporates a hierarchical attentive heterogeneous graph network in Albert~\cite{lan2019albert} to perform redundancy-aware sentence embedding for extractive summarization.
\textit{Point-Generator}~\cite{see2017get} first introduces the copy mechanism with coverage regularization for attentive seq2seq models in abstractive summarization. 
\textit{BART}~\cite{lewis2020bart} is an encoder-decoder Transformer model with denoising seq2seq pre-training.
\textit{PEGASUS}~\cite{zhang2020pegasus} uses gap-sentence generation for pre-training an encoder-decoder Transformer on abstractive summarization tasks.
\textit{CIT+SE}~\cite{saito2020abstractive} uses a key word extractor and selective encoding mechanism to guide abstractive summarization.
\textit{GSum}~\cite{dou2021gsum} uses the extracted summary from MatchSum as guidance to supervise BART for abstractive summarization.

\stitle{Implementation details}
We fine-tune BART-large on CNNDM and Newsroom datasets.
For training data, we prepend a special token in front of each sentence for calculating its sentence representation.
Each input sequence is truncated to 1024 tokens (including special tokens), to fit the maximum input length for BART.
According to the summary length distribution of CNNDM and Newsroom datasets, we truncated the reference summary to be 128 and 256 tokens respectively to ensure more than 99\% of the reference summaries are fully preserved.
For inference, we use the predicted soft estimation for allocation of expected salience.
The predicted probability of salience degree is sharpened with a temperature $\tau=0.5$.
We use beam search with beam size of 5, length penalty of 1.5 and 3-gram blocking.
According to their ROGUE-L $F_1$ scores against the ground-truth summary, we split sentences into $L=3$ categories of salience degrees: 1) The most important top 15\% sentences, 2) the bottom 50\% least important ones, and 3) everything in between. More discussions regarding this setup is in \Cref{sec:principle}).

\subsection{Main Results}
\label{sec:main_result}
\TABLEMAIN
\Cref{tab:main} shows the results on the two summarization datasets.
For baselines on CNNDM, joint extractive and abstractive summarization methods (i.e. CIT+SE and GSum) perform better than independent extractive %
(i.e. LEAD-3, MatchSum and HAHSum) and abstractive summarization methods (i.e. Point-Generator, BART and PEGASUS) when using the same backbone models. 
Among the joint summarization baselines, using extractive summary as guidance (i.e. GSum) performs better than using key words as guidance (i.e. CIT+SE), which agrees with the observation by \citet{dou2021gsum}. 
Our method %
promisingly improves the
original BART by 2.06/1.41/1.91 points in terms of ROUGE-1/2/L F1 scores, indicating the multi-degree salience expectation allocation effectively guides the model to generate better-quality summaries.
In comparison with GSum, our method achieves improvements of 0.33/0.32/0.60 points in terms of ROUGE-1/2/L F1 scores, w/o the help of SOTA extractive summarization system and with less additional parameters, indicating the proposed guidance is more effective than extractive summaries.
Results on Newsroom dataset further verify that our method can achieve consistent improvements on different datasets. In comparison with the vanilla BART model, our method achieves 0.50/0.32/0.34 points improvements in terms of ROUGE-1/2/L F1 scores.

\subsection{The Fifteen-Fifty Phenomenon}
\label{sec:principle}
\TABLESALIENCETHRESHOLDCNNDM
\FIGURECNNDMSUBSETS

The number of salience degrees and thresholds to delimit them are the important hyper-parameters to discretize the proposed guidance.
We apply a greedy search algorithm to find the best thresholds.
First, we compute salience scores of all sentences in the corpus. In this work, we use ROUGE-L F$_{1}$ between each document sentence and corresponding reference summary to represent salience, and find the best threshold for two salience degrees.
Then we gradually add one more salience degree and search the additional threshold.
The results on CNNDM is shown in \Cref{tab:threshold}.
Splitting all sentences into three salience degrees by top 15\% and bottom 50\% salience scores leads to the best ROUGE-L F$_{1}$.
As the number of salience degrees $L$ increases, the model performance first increases and then decreases. We attribute this phenomenon to the trade-off between informativeness and prediction accuracy of the guidance. Although a more fine-grained salience guidance is more informative, our model generates summaries based on predicted guidance during inference, where error propagation exists. Increasing the number of salience degrees to predict also increases the risk of misclassification.
Furthermore, we find the best number and thresholds of salience degrees for summarization is consistent on Newsroom, indicating the salience split by top fifteen and bottom fifty percentile is a nature property of news articles.
This phenomenon may provide a useful insight for composing news articles by journalists.

\TABLEHUMANEVAL
\TABLEHUMANRANK

\section{Analysis}
To gain further insights on the proposed method, we perform additional analyses on CNNDM to comprehensively investigate performance by abstractiveness (\Cref{sec:abstractiveness}), 
summary length (\Cref{sec:length}), human evaluation (\Cref{sec:human}), and the impact of different model components (\Cref{sec:ablation}).  A case study is also presented in \Cref{sec:case}.

\subsection{Performance by Abstractiveness}
\label{sec:abstractiveness}
To understand %
how adaptive our method is on documents with different abstractiveness, we split all documents into three subsets of equal sizes based on their density scores following \citet{grusky2018newsroom}. Results are shown in \Cref{fig:cnndm_subsets}. \MODEL performs better than baselines on all subsets, indicating that our method is adaptive to documents with different abstractiveness. The improvements on \textit{abstractive} and \textit{mixed} subsets are slightly higher than that on the \textit{extractive} subset, indicating abstract documents benefit more than extractive ones from a flexible salience guidance.

\subsection{Summary Length}
\label{sec:length}
As \MODEL achieves better performance under different abstractiveness with a more flexible salience guidance, a followup research question is: \textit{Does the flexible salience guidance help predict summary length more accurately?}
To answer this question, we compute the average lengths of Ground-Truth summaries and summaries generated by \MODEL and baseline systems.
The average summary length of Ground-Truth, SEASON, BART, and GSum are respectively 54.8, 59.0, 60.7, and 72.0.
Among these methods, SEASON gave the closest average summary length to Ground-Truth.
Moreover, while both of \MODEL and GSum introduce sentence-level salience guidance to BART, they change the summary length in opposite directions.

\subsection{Human Evaluation}
\label{sec:human}
\TABLESALIENCEINJECTION
\TABLEAUXILIARYTASK

We further evaluate the system outputs of \MODEL, BART, Gsum and the Ground-Truth with human subjective evaluation.
We randomly pick 100 instances from CNNDM test set.
For each raw document in those instances, we provide the summary generated by each system and the ground-truth summary.
We hire human evaluators on Amazon Mechanical Turk to answer three Yes/No questions for the four summaries and rank them.
Each instance is assigned to 3 different human evaluators to answer the following
three questions~\cite{song-etal-2021-new}. 
a) \textit{Informativeness}: Does the summary include the major information from the news? 
b) \textit{Faithfulness}: Does the summary give any additional information not covered by the news? 
c) \textit{Fluency}: Is the summary grammatical and well-formed?

\Cref{tab:human_evaluation_questions} %
reports the average percentage for each method to get a positive answer on the corresponding question. 
Among the three systems, \MODEL performs the best on informativeness and fluency, while GSum performs the best on faithfulness.
This indicates that our flexible guidance helps the model to identify salient content accurately and rephrase them properly.
Not surprisingly, systems with guidance (i.e. \MODEL and Gsum) are more faithful to original content than a system without any guidance (i.e. BART).
\Cref{tab:human_evaluation_ranking} shows the ranking results.
\MODEL has the highest percentage of the highest-ranked summaries and the lowest percentage of the lowest-ranked summaries.
It also has the best average rank.
These results further demonstrate that summaries generated by \MODEL are of high quality.
Interestingly, we find that the ground-truth is not always the best choice in human evaluation.
This observation aligns with the findings in prior studies \cite{maynez2020faithfulness,song2020controlling,fabbri2021summeval}.
It could happen since both human composition of summaries and human justification on their qualities %
could be subjective.
Thus, ground-truth news summaries written by editors may not always be the first choice of readers.
It also indicates that human evaluators could not distinguish between the real human writer and our automatic summarizer, and actually prefer our system outputs more than the ground-truth summaries.

\subsection{Ablation Study}
\label{sec:ablation}
\TABLELABELSMOOTHING
\TABLEMARGINALIZEDDECODING
\TABLECASE

For all the experiments in this section, we use the default setting introduced in \Cref{sec:setup}, unless discussed otherwise with different hyper-parameter values.

\stitle{Multi-Task Learning}
We first investigate the effectiveness of salience prediction as an auxiliary task by removing salience-aware cross-attention.
In this setting, the model jointly predicts the salience allocation and the abstractive summary, but does not feed the gold or predicted salience allocation to the decoder. 
That means the salience allocation is only used as supervision signals but not (intermediate) input features.
As shown in \Cref{tab:ablation_module}, with MTL solely, the model can achieve 0.36/0.32/0.32 points improvements in terms of R-1/2/L. This indicates that the salience prediction task can not only provide effective guidance for abstractive summarization, but also act as supervision for learning more robust representations.

\stitle{Salience-Aware Cross-Attention}
We examine the effectiveness of the proposed salience-aware cross-attention module from two perspectives in \Cref{tab:ablation_module}.
First, we provide the gold salience labels instead of predicted ones to explore its upper bound. The performance increases by 8.58/8.72/9.06 points with a perfect salience predictor. This result indicates that a better estimation of the salience can be helpful for further improving the abstractive summarization performance.
Second, we compare it with the original cross-attention module while keeping the auxiliary task and observe a performance drop by 1.70/1.09/1.59 points in terms of R-1/2/L. This indicates that salience-aware cross-attention is essential for selecting important content accurately.

\stitle{Coefficient of Multi-Task Learning}
We further examine the influence of the coefficient $\alpha$ of multi-task learning in \Cref{tab:coefficient}. We test three different $\alpha$ values and observe that the largest difference of R1/2/L are within 0.02/0.07/0.03. According to the results, \MODEL is not sensitive to $\alpha$, indicating our model architecture is robust.

\stitle{Adjacent Label Smoothing}
We compare different label smoothing strategies in \Cref{tab:label_smoothing}. In general, label smoothing improves model generalization and calibration~\cite{muller2019does}, therefore benefits the overall performance.
Given the same smoothing probability $\beta$, adding label smoothing to adjacent salience degrees performs better than adding label smoothing on all other salient degrees.

\stitle{Salience Estimation}
We compare the effectiveness of using soft (\Cref{eq:function_f_soft}) and hard (\Cref{eq:function_f_hard}) strategies for salience estimation in \Cref{tab:salience_expection}. Computing the expectation with raw probabilities (i.e., $\tau=1.0$) brings 0.11 points improvements on ROUGE-1. By adjusting the sharpness of probability distribution with the sharpening coefficient $\tau$, ROUGE-1/2/L improvements become 0.50/0.39/0.34 points, respectively. As defined in \Cref{eq:cls_prob}, $\tau$ represents the confidence on predictions, and a lower $\tau$ leads to sharper probability distribution. In our experiments, $\tau=0.5$ performs the best.

\subsection{Case Study}
\label{sec:case}

We present a case study in \Cref{tab:case} with two representative examples to illustrate the advantage of \MODEL. 
In the first case, 
BART tends to generate extra details without the help of proper guidance when only one sentence is enough to summarize the document. 
GSum is guided by an extractive summary consisting of three sentences, so not surprisingly it provides even more details. 
In the second case, 
BART infers without any salience guidance and ignores an important finding of the research. GSum selects exactly three sentences as guidance, thus it misses key information when multiple sentences are similarly important but some of them are not included in the guidance.
\MODEL performs well for both cases, indicating it is adaptive to documents of different properties.

\section{Conclusion}

In this paper, we propose \MODEL, an abstractive summarization approach guided with salience allocation expectation. In \MODEL, the salience guidance is adaptive to documents with different abstractiveness, and the salience-aware cross-attention module is flexible to decide how much signal to accept from the salience guidance.
Automatic and human evaluation further demonstrate the effectiveness and reliability of our proposed method.
Comparing to the strong baseline model (i.e. BART), our method achieves 2.06/1.41/1.91 ROUGE-1/2/L performance gain on CNNDM, and 0.33/0.32/0.60 performance gain on Newsroom.
Finally, the empirical results on more than one million news articles demonstrate a natural fifteen-fifty salience split for news article sentences providing a useful insight for composing news articles.

\section*{Ackonwledgement}
We thank all reviewers for their valuable suggestions.
We also thank Zi-Yi Dou and Pengfei Liu for providing summaries generated by GSum and stored at DataLab~\cite{xiao-etal-2022-datalab}\footnote{\url{https://datalab.nlpedia.ai/}}.
This work was done when Fei Wang was doing an internship at Tencent AI Lab Seattle.
Fei Wang is partially supported by Annerberg Fellowship.
Muhao Chen is supported by the National Science Foundation of United States Grant IIS 2105329.

\section*{Limitation}

In this study we have experimented with using ROUGE-L F$_{1}$ as the salience measurement. However, other metrics for text summarization can serve as alternatives, such as BLEU \cite{papineni2002bleu} and BARTScore \cite{yuan2021bartscore}. Choosing among more metrics for salience measurement can be explored in future work.
The number of salience degrees and thresholds to delimit them could be language-dependent. The fifty-fifteen phenomenon is observed on two of the most representative English news summarization datasets. Future work on other languages may need to search the best salience degrees and thresholds of each language.
Despite we use BART as our base model and maximum likelihood estimation (MLE) as the learning objective in this study, the proposed method can also be applied to other backbones and learning objectives \cite{zhang2020pegasus,liu2021simcls,liu2022brio}.
While we have limited the proposed technique to abstractive summarization on news articles, future research can extend \MODEL to other domains, such as scientific publications \cite{cohan2018discourse} and podcasts \cite{song2022towards}.
In terms of evaluation, we focus on the supervised and in-domain setting. Future work may also consider to extend our method to zero-shot, few-shot, cross-domain, or cross-dataset settings.
In addition to abstractive summarization, future research can also extend \MODEL to other NLP tasks requiring salience-awareness, such as fact verification \cite{wang2021table}, information retrieval \cite{xiong2018towards} and distantly supervised relation extraction \cite{lin2016neural}.

\section*{Ethical Consideration}
A general issue of automatic text summarization is intellectual property problem caused by copying content from the raw document to the generated summary. 
This work seeks to improve abstractive summarization models with salience allocation as guidance.
As the proposed guidance is more flexible than extractive summaries, it is likely to reduce copying content.
Although we create salience guidance based on ground-truth summaries, the documents and ground-truth summaries remains the same as it is in the original dataset, ensuring no further social bias is introduced.

\bibliography{anthology,ref}

\begin{thebibliography}{51}
\expandafter\ifx\csname natexlab\endcsname\relax\def\natexlab#1{#1}\fi

\bibitem[{Beltagy et~al.(2020)Beltagy, Peters, and
  Cohan}]{Beltagy2020Longformer}
Iz~Beltagy, Matthew~E. Peters, and Arman Cohan. 2020.
\newblock Longformer: The long-document transformer.
\newblock \emph{arXiv:2004.05150}.

\bibitem[{Cheng et~al.(2015)Cheng, Xu, and Qu}]{cheng2015summarizing}
Gong Cheng, Danyun Xu, and Yuzhong Qu. 2015.
\newblock Summarizing entity descriptions for effective and efficient
  human-centered entity linking.
\newblock In \emph{Proceedings of the 24th International Conference on World
  Wide Web}, pages 184--194.

\bibitem[{Cohan et~al.(2018)Cohan, Dernoncourt, Kim, Bui, Kim, Chang, and
  Goharian}]{cohan2018discourse}
Arman Cohan, Franck Dernoncourt, Doo~Soon Kim, Trung Bui, Seokhwan Kim, Walter
  Chang, and Nazli Goharian. 2018.
\newblock A discourse-aware attention model for abstractive summarization of
  long documents.
\newblock In \emph{Proceedings of the 2018 Conference of the North American
  Chapter of the Association for Computational Linguistics: Human Language
  Technologies, Volume 2 (Short Papers)}, pages 615--621.

\bibitem[{Diaz and Marathe(2019)}]{diaz2019soft}
Raul Diaz and Amit Marathe. 2019.
\newblock Soft labels for ordinal regression.
\newblock In \emph{Proceedings of the IEEE/CVF conference on computer vision
  and pattern recognition}, pages 4738--4747.

\bibitem[{Dou et~al.(2021)Dou, Liu, Hayashi, Jiang, and Neubig}]{dou2021gsum}
Zi-Yi Dou, Pengfei Liu, Hiroaki Hayashi, Zhengbao Jiang, and Graham Neubig.
  2021.
\newblock Gsum: A general framework for guided neural abstractive
  summarization.
\newblock In \emph{Proceedings of the 2021 Conference of the North American
  Chapter of the Association for Computational Linguistics: Human Language
  Technologies}, pages 4830--4842.

\bibitem[{Fabbri et~al.(2021)Fabbri, Kry{\'s}ci{\'n}ski, McCann, Xiong, Socher,
  and Radev}]{fabbri2021summeval}
Alexander~Richard Fabbri, Wojciech Kry{\'s}ci{\'n}ski, Bryan McCann, Caiming
  Xiong, Richard Socher, and Dragomir Radev. 2021.
\newblock Summeval: Re-evaluating summarization evaluation.
\newblock \emph{Transactions of the Association for Computational Linguistics},
  9:391--409.

\bibitem[{Gehrmann et~al.(2018)Gehrmann, Deng, and Rush}]{gehrmann2018bottom}
Sebastian Gehrmann, Yuntian Deng, and Alexander~M Rush. 2018.
\newblock Bottom-up abstractive summarization.
\newblock In \emph{Proceedings of the 2018 Conference on Empirical Methods in
  Natural Language Processing}, pages 4098--4109.

\bibitem[{Grusky et~al.(2018)Grusky, Naaman, and Artzi}]{grusky2018newsroom}
Max Grusky, Mor Naaman, and Yoav Artzi. 2018.
\newblock Newsroom: A dataset of 1.3 million summaries with diverse extractive
  strategies.
\newblock In \emph{Proceedings of the 2018 Conference of the North American
  Chapter of the Association for Computational Linguistics: Human Language
  Technologies, Volume 1 (Long Papers)}, pages 708--719.

\bibitem[{Hsu et~al.(2018)Hsu, Lin, Lee, Min, Tang, and Sun}]{hsu2018unified}
Wan-Ting Hsu, Chieh-Kai Lin, Ming-Ying Lee, Kerui Min, Jing Tang, and Min Sun.
  2018.
\newblock A unified model for extractive and abstractive summarization using
  inconsistency loss.
\newblock In \emph{Proceedings of the 56th Annual Meeting of the Association
  for Computational Linguistics (Volume 1: Long Papers)}, pages 132--141.

\bibitem[{Inoue et~al.(2021)Inoue, Trivedi, Sinha, Balasubramanian, and
  Inui}]{inoue-etal-2021-summarize}
Naoya Inoue, Harsh Trivedi, Steven Sinha, Niranjan Balasubramanian, and Kentaro
  Inui. 2021.
\newblock \href {https://doi.org/10.18653/v1/2021.emnlp-main.490}
  {Summarize-then-answer: Generating concise explanations for multi-hop reading
  comprehension}.
\newblock In \emph{Proceedings of the 2021 Conference on Empirical Methods in
  Natural Language Processing}, pages 6064--6080, Online and Punta Cana,
  Dominican Republic. Association for Computational Linguistics.

\bibitem[{Jia et~al.(2020)Jia, Cao, Tang, Fang, Cao, and Wang}]{jia2020neural}
Ruipeng Jia, Yanan Cao, Hengzhu Tang, Fang Fang, Cong Cao, and Shi Wang. 2020.
\newblock Neural extractive summarization with hierarchical attentive
  heterogeneous graph network.
\newblock In \emph{Proceedings of the 2020 Conference on Empirical Methods in
  Natural Language Processing (EMNLP)}, pages 3622--3631.

\bibitem[{Jin et~al.(2020)Jin, Wang, and Wan}]{jin2020semsum}
Hanqi Jin, Tianming Wang, and Xiaojun Wan. 2020.
\newblock Semsum: Semantic dependency guided neural abstractive summarization.
\newblock In \emph{Proceedings of the AAAI Conference on Artificial
  Intelligence}, volume~34, pages 8026--8033.

\bibitem[{Johnston and Dark(1986)}]{johnston1986selective}
William~A Johnston and Veronica~J Dark. 1986.
\newblock Selective attention.
\newblock \emph{Annual review of psychology}.

\bibitem[{Koupaee and Wang(2018)}]{koupaee2018wikihow}
Mahnaz Koupaee and William~Yang Wang. 2018.
\newblock Wikihow: A large scale text summarization dataset.
\newblock \emph{arXiv preprint arXiv:1810.09305}.

\bibitem[{Lan et~al.(2019)Lan, Chen, Goodman, Gimpel, Sharma, and
  Soricut}]{lan2019albert}
Zhenzhong Lan, Mingda Chen, Sebastian Goodman, Kevin Gimpel, Piyush Sharma, and
  Radu Soricut. 2019.
\newblock Albert: A lite bert for self-supervised learning of language
  representations.
\newblock In \emph{International Conference on Learning Representations}.

\bibitem[{Lewis et~al.(2020)Lewis, Liu, Goyal, Ghazvininejad, Mohamed, Levy,
  Stoyanov, and Zettlemoyer}]{lewis2020bart}
Mike Lewis, Yinhan Liu, Naman Goyal, Marjan Ghazvininejad, Abdelrahman Mohamed,
  Omer Levy, Veselin Stoyanov, and Luke Zettlemoyer. 2020.
\newblock Bart: Denoising sequence-to-sequence pre-training for natural
  language generation, translation, and comprehension.
\newblock In \emph{Proceedings of the 58th Annual Meeting of the Association
  for Computational Linguistics}, pages 7871--7880.

\bibitem[{Li et~al.(2021)Li, Einolghozati, Iyer, Paranjape, Mehdad, Gupta, and
  Ghazvininejad}]{li2021ease}
Haoran Li, Arash Einolghozati, Srinivasan Iyer, Bhargavi Paranjape, Yashar
  Mehdad, Sonal Gupta, and Marjan Ghazvininejad. 2021.
\newblock Ease: Extractive-abstractive summarization end-to-end using the
  information bottleneck principle.
\newblock In \emph{Proceedings of the Third Workshop on New Frontiers in
  Summarization}, pages 85--95.

\bibitem[{Li et~al.(2020)Li, Zhu, Zhang, Zong, and He}]{li2020keywords}
Haoran Li, Junnan Zhu, Jiajun Zhang, Chengqing Zong, and Xiaodong He. 2020.
\newblock Keywords-guided abstractive sentence summarization.
\newblock In \emph{Proceedings of the AAAI Conference on Artificial
  Intelligence}, volume~34, pages 8196--8203.

\bibitem[{Lin(2004)}]{lin2004rouge}
Chin-Yew Lin. 2004.
\newblock Rouge: A package for automatic evaluation of summaries.
\newblock In \emph{Text summarization branches out}, pages 74--81.

\bibitem[{Lin et~al.(2016)Lin, Shen, Liu, Luan, and Sun}]{lin2016neural}
Yankai Lin, Shiqi Shen, Zhiyuan Liu, Huanbo Luan, and Maosong Sun. 2016.
\newblock Neural relation extraction with selective attention over instances.
\newblock In \emph{Proceedings of the 54th Annual Meeting of the Association
  for Computational Linguistics (Volume 1: Long Papers)}, pages 2124--2133.

\bibitem[{Liu and Liu(2009)}]{liu2009extractive}
Fei Liu and Yang Liu. 2009.
\newblock From extractive to abstractive meeting summaries: Can it be done by
  sentence compression?
\newblock In \emph{Proceedings of the ACL-IJCNLP 2009 Conference Short Papers},
  pages 261--264.

\bibitem[{Liu(2019)}]{liu2019fine}
Yang Liu. 2019.
\newblock Fine-tune bert for extractive summarization.
\newblock \emph{arXiv preprint arXiv:1903.10318}.

\bibitem[{Liu and Liu(2021)}]{liu2021simcls}
Yixin Liu and Pengfei Liu. 2021.
\newblock Simcls: A simple framework for contrastive learning of abstractive
  summarization.
\newblock In \emph{Proceedings of the 59th Annual Meeting of the Association
  for Computational Linguistics and the 11th International Joint Conference on
  Natural Language Processing (Volume 2: Short Papers)}, pages 1065--1072.

\bibitem[{Liu et~al.(2022)Liu, Liu, Radev, and Neubig}]{liu2022brio}
Yixin Liu, Pengfei Liu, Dragomir Radev, and Graham Neubig. 2022.
\newblock Brio: Bringing order to abstractive summarization.
\newblock In \emph{Proceedings of the 60th Annual Meeting of the Association
  for Computational Linguistics (Volume 1: Long Papers)}, pages 2890--2903.

\bibitem[{Loshchilov and Hutter(2018)}]{loshchilov2018decoupled}
Ilya Loshchilov and Frank Hutter. 2018.
\newblock Decoupled weight decay regularization.
\newblock In \emph{International Conference on Learning Representations}.

\bibitem[{Lu et~al.(2022)Lu, Hsu, Zhou, Ma, Chen et~al.}]{lu2022summarization}
Keming Lu, I~Hsu, Wenxuan Zhou, Mingyu~Derek Ma, Muhao Chen, et~al. 2022.
\newblock Summarization as indirect supervision for relation extraction.
\newblock \emph{arXiv preprint arXiv:2205.09837}.

\bibitem[{Manakul and Gales(2021)}]{manakul2021long}
Potsawee Manakul and Mark Gales. 2021.
\newblock Long-span summarization via local attention and content selection.
\newblock In \emph{Proceedings of the 59th Annual Meeting of the Association
  for Computational Linguistics and the 11th International Joint Conference on
  Natural Language Processing (Volume 1: Long Papers)}, pages 6026--6041.

\bibitem[{Maynez et~al.(2020)Maynez, Narayan, Bohnet, and
  McDonald}]{maynez2020faithfulness}
Joshua Maynez, Shashi Narayan, Bernd Bohnet, and Ryan McDonald. 2020.
\newblock On faithfulness and factuality in abstractive summarization.
\newblock In \emph{Proceedings of the 58th Annual Meeting of the Association
  for Computational Linguistics}, pages 1906--1919.

\bibitem[{M{\"u}ller et~al.(2019)M{\"u}ller, Kornblith, and
  Hinton}]{muller2019does}
Rafael M{\"u}ller, Simon Kornblith, and Geoffrey~E Hinton. 2019.
\newblock When does label smoothing help?
\newblock \emph{Advances in neural information processing systems}, 32.

\bibitem[{Nallapati et~al.(2016)Nallapati, Zhou, dos Santos,
  G{\.{u}}l{\c{c}}ehre, and Xiang}]{nallapati2016abstractive}
Ramesh Nallapati, Bowen Zhou, Cicero dos Santos, {\c{C}}a{\u{g}}lar
  G{\.{u}}l{\c{c}}ehre, and Bing Xiang. 2016.
\newblock Abstractive text summarization using sequence-to-sequence rnns and
  beyond.
\newblock In \emph{Proceedings of The 20th SIGNLL Conference on Computational
  Natural Language Learning}, pages 280--290.

\bibitem[{Papineni et~al.(2002)Papineni, Roukos, Ward, and
  Zhu}]{papineni2002bleu}
Kishore Papineni, Salim Roukos, Todd Ward, and Wei-Jing Zhu. 2002.
\newblock Bleu: a method for automatic evaluation of machine translation.
\newblock In \emph{Proceedings of the 40th annual meeting of the Association
  for Computational Linguistics}, pages 311--318.

\bibitem[{Pilault et~al.(2020)Pilault, Li, Subramanian, and
  Pal}]{pilault2020extractive}
Jonathan Pilault, Raymond Li, Sandeep Subramanian, and Christopher Pal. 2020.
\newblock On extractive and abstractive neural document summarization with
  transformer language models.
\newblock In \emph{Proceedings of the 2020 Conference on Empirical Methods in
  Natural Language Processing (EMNLP)}, pages 9308--9319.

\bibitem[{Pruthi et~al.(2020)Pruthi, Gupta, Dhingra, Neubig, and
  Lipton}]{pruthi2020learning}
Danish Pruthi, Mansi Gupta, Bhuwan Dhingra, Graham Neubig, and Zachary~C
  Lipton. 2020.
\newblock Learning to deceive with attention-based explanations.
\newblock In \emph{Proceedings of the 58th Annual Meeting of the Association
  for Computational Linguistics}, pages 4782--4793.

\bibitem[{Rasley et~al.(2020)Rasley, Rajbhandari, Ruwase, and
  He}]{rasley2020deepspeed}
Jeff Rasley, Samyam Rajbhandari, Olatunji Ruwase, and Yuxiong He. 2020.
\newblock Deepspeed: System optimizations enable training deep learning models
  with over 100 billion parameters.
\newblock In \emph{Proceedings of the 26th ACM SIGKDD International Conference
  on Knowledge Discovery \& Data Mining}, pages 3505--3506.

\bibitem[{Rush et~al.(2015)Rush, Chopra, and Weston}]{rush2015neural}
Alexander~M Rush, Sumit Chopra, and Jason Weston. 2015.
\newblock A neural attention model for abstractive sentence summarization.
\newblock In \emph{Proceedings of the 2015 Conference on Empirical Methods in
  Natural Language Processing}, pages 379--389.

\bibitem[{Saito et~al.(2020)Saito, Nishida, Nishida, and
  Tomita}]{saito2020abstractive}
Itsumi Saito, Kyosuke Nishida, Kosuke Nishida, and Junji Tomita. 2020.
\newblock Abstractive summarization with combination of pre-trained
  sequence-to-sequence and saliency models.
\newblock \emph{arXiv preprint arXiv:2003.13028}.

\bibitem[{See et~al.(2017)See, Liu, and Manning}]{see2017get}
Abigail See, Peter~J Liu, and Christopher~D Manning. 2017.
\newblock Get to the point: Summarization with pointer-generator networks.
\newblock In \emph{Proceedings of the 55th Annual Meeting of the Association
  for Computational Linguistics (Volume 1: Long Papers)}, pages 1073--1083.

\bibitem[{Song et~al.(2022)Song, Li, Wang, Yu, and Liu}]{song2022towards}
Kaiqiang Song, Chen Li, Xiaoyang Wang, Dong Yu, and Fei Liu. 2022.
\newblock Towards abstractive grounded summarization of podcast transcripts.
\newblock In \emph{Proceedings of the 60th Annual Meeting of the Association
  for Computational Linguistics (Volume 1: Long Papers)}, pages 4407--4418.

\bibitem[{Song et~al.(2021)Song, Wang, Feng, and Liu}]{song-etal-2021-new}
Kaiqiang Song, Bingqing Wang, Zhe Feng, and Fei Liu. 2021.
\newblock \href {https://doi.org/10.18653/v1/2021.naacl-main.110} {A new
  approach to overgenerating and scoring abstractive summaries}.
\newblock In \emph{Proceedings of the 2021 Conference of the North American
  Chapter of the Association for Computational Linguistics: Human Language
  Technologies}, pages 1392--1404, Online. Association for Computational
  Linguistics.

\bibitem[{Song et~al.(2020)Song, Wang, Feng, Liu, and
  Liu}]{song2020controlling}
Kaiqiang Song, Bingqing Wang, Zhe Feng, Ren Liu, and Fei Liu. 2020.
\newblock Controlling the amount of verbatim copying in abstractive
  summarization.
\newblock In \emph{Proceedings of the AAAI Conference on Artificial
  Intelligence}, volume~34, pages 8902--8909.

\bibitem[{Sukhbaatar et~al.(2019)Sukhbaatar, Grave, Bojanowski, and
  Joulin}]{sukhbaatar2019adaptive}
Sainbayar Sukhbaatar, {\'E}douard Grave, Piotr Bojanowski, and Armand Joulin.
  2019.
\newblock Adaptive attention span in transformers.
\newblock In \emph{Proceedings of the 57th Annual Meeting of the Association
  for Computational Linguistics}, pages 331--335.

\bibitem[{Wang et~al.(2021)Wang, Sun, Pujara, Szekely, and
  Chen}]{wang2021table}
Fei Wang, Kexuan Sun, Jay Pujara, Pedro Szekely, and Muhao Chen. 2021.
\newblock Table-based fact verification with salience-aware learning.
\newblock In \emph{Findings of the Association for Computational Linguistics:
  EMNLP 2021}, pages 4025--4036.

\bibitem[{Wang et~al.(2022)Wang, Xu, Szekely, and Chen}]{wang2022robust}
Fei Wang, Zhewei Xu, Pedro Szekely, and Muhao Chen. 2022.
\newblock Robust (controlled) table-to-text generation with structure-aware
  equivariance learning.
\newblock \emph{arXiv preprint arXiv:2205.03972}.

\bibitem[{Wolf et~al.(2020)Wolf, Debut, Sanh, Chaumond, Delangue, Moi, Cistac,
  Rault, Louf, Funtowicz, Davison, Shleifer, von Platen, Ma, Jernite, Plu, Xu,
  Le~Scao, Gugger, Drame, Lhoest, and Rush}]{wolf-etal-2020-transformers}
Thomas Wolf, Lysandre Debut, Victor Sanh, Julien Chaumond, Clement Delangue,
  Anthony Moi, Pierric Cistac, Tim Rault, Remi Louf, Morgan Funtowicz, Joe
  Davison, Sam Shleifer, Patrick von Platen, Clara Ma, Yacine Jernite, Julien
  Plu, Canwen Xu, Teven Le~Scao, Sylvain Gugger, Mariama Drame, Quentin Lhoest,
  and Alexander Rush. 2020.
\newblock \href {https://doi.org/10.18653/v1/2020.emnlp-demos.6} {Transformers:
  State-of-the-art natural language processing}.
\newblock In \emph{Proceedings of the 2020 Conference on Empirical Methods in
  Natural Language Processing: System Demonstrations}, pages 38--45, Online.
  Association for Computational Linguistics.

\bibitem[{Xiao et~al.(2022)Xiao, Fu, Yuan, Viswanathan, Liu, Liu, Neubig, and
  Liu}]{xiao-etal-2022-datalab}
Yang Xiao, Jinlan Fu, Weizhe Yuan, Vijay Viswanathan, Zhoumianze Liu, Yixin
  Liu, Graham Neubig, and Pengfei Liu. 2022.
\newblock \href {https://doi.org/10.18653/v1/2022.acl-demo.18} {{D}ata{L}ab: A
  platform for data analysis and intervention}.
\newblock In \emph{Proceedings of the 60th Annual Meeting of the Association
  for Computational Linguistics: System Demonstrations}, pages 182--195,
  Dublin, Ireland. Association for Computational Linguistics.

\bibitem[{Xiong et~al.(2018)Xiong, Liu, Callan, and Liu}]{xiong2018towards}
Chenyan Xiong, Zhengzhong Liu, Jamie Callan, and Tie-Yan Liu. 2018.
\newblock Towards better text understanding and retrieval through kernel entity
  salience modeling.
\newblock In \emph{The 41st International ACM SIGIR Conference on Research \&
  Development in Information Retrieval}, pages 575--584.

\bibitem[{Yin et~al.(2021)Yin, Radev, and Xiong}]{yin-etal-2021-docnli}
Wenpeng Yin, Dragomir Radev, and Caiming Xiong. 2021.
\newblock \href {https://doi.org/10.18653/v1/2021.findings-acl.435}
  {{D}oc{NLI}: A large-scale dataset for document-level natural language
  inference}.
\newblock In \emph{Findings of the Association for Computational Linguistics:
  ACL-IJCNLP 2021}, pages 4913--4922, Online. Association for Computational
  Linguistics.

\bibitem[{Yuan et~al.(2021)Yuan, Neubig, and Liu}]{yuan2021bartscore}
Weizhe Yuan, Graham Neubig, and Pengfei Liu. 2021.
\newblock Bartscore: Evaluating generated text as text generation.
\newblock \emph{Advances in Neural Information Processing Systems},
  34:27263--27277.

\bibitem[{Zhang et~al.(2020)Zhang, Zhao, Saleh, and Liu}]{zhang2020pegasus}
Jingqing Zhang, Yao Zhao, Mohammad Saleh, and Peter Liu. 2020.
\newblock Pegasus: Pre-training with extracted gap-sentences for abstractive
  summarization.
\newblock In \emph{International Conference on Machine Learning}, pages
  11328--11339. PMLR.

\bibitem[{Zhong et~al.(2020)Zhong, Liu, Chen, Wang, Qiu, and
  Huang}]{zhong2020extractive}
Ming Zhong, Pengfei Liu, Yiran Chen, Danqing Wang, Xipeng Qiu, and Xuan-Jing
  Huang. 2020.
\newblock Extractive summarization as text matching.
\newblock In \emph{Proceedings of the 58th Annual Meeting of the Association
  for Computational Linguistics}, pages 6197--6208.

\bibitem[{Zhong et~al.(2019)Zhong, Liu, Wang, Qiu, and
  Huang}]{zhong-etal-2019-searching}
Ming Zhong, Pengfei Liu, Danqing Wang, Xipeng Qiu, and Xuanjing Huang. 2019.
\newblock \href {https://doi.org/10.18653/v1/P19-1100} {Searching for effective
  neural extractive summarization: What works and what{'}s next}.
\newblock In \emph{Proceedings of the 57th Annual Meeting of the Association
  for Computational Linguistics}, pages 1049--1058, Florence, Italy.
  Association for Computational Linguistics.

\end{thebibliography}
\bibliographystyle{style/acl_natbib}

\clearpage
\appendix
\section{Implementation Details}
\label{sec:appendix_implementation}
Our Implementation is based on Huggingface Transformers~\cite{wolf-etal-2020-transformers}, Pytorch-lightning\footnote{\url{https://www.pytorchlightning.ai}} and Lightning-Transformers\footnote{\url{https://github.com/Lightning-AI/lightning-transformers}}. We further use DeepSpeed \cite{rasley2020deepspeed} Stage2 and half-precision to speed up the training process.
We applied the BART-large model consisting of 400M parameters and fine-tuned them on CNNDM\footnote{\url{https://huggingface.co/datasets/ccdv/cnn_dailymail}} and Newsroom\footnote{\url{https://lil.nlp.cornell.edu/newsroom/download/index.html}} dataset with 8$\times$V100 GPU(32GB) for 10 epochs. It takes about 5 hours for training on CNNDM and 32 hours on NEWSROOM.
We set our batch size to be 96 to maximize the utilization of the GPU memory.
For Optimization, we use AdamW \cite{loshchilov2018decoupled} with learning rates of $3e-5$.
The momentum parameters are 0.9 and 0.99.
Our weight decay is set to be $0.01$ for parameters other than bias term and LayerNorm layers.
We uses a linear warmup strategy, the number of warmup steps are 1,500 and 5,000 on CNNDM and Newsroom respectively.
The dropout rate is 10\%.
We follow the BART \cite{lewis2020bart} fine-tuning and use gradient clipping of 0.1.
The coefficient of multi-task learning $\alpha$ is 1.5.
We uses adjacent label smoothing and the smoothing probability $\beta$ is 20\%.
For inference, we use the predicted soft estimation for allocation of expected salience.
The predicted probability of salience degree is sharpened with a temperature $\tau=0.5$.
We use beam search with beam size of 5 with length penalty of 1.5 and 3-gram blocking.
The minimum and maximum decoding length is 20 and 256.
For human evaluation, we choose master workers on Amazon Mechanical Turk\footnote{\url{https://www.mturk.com}} with
more than 90\% approve rates and more than 100 approved HIT.

\end{document}